\documentclass[runningheads]{llncs}
\usepackage{graphicx}
\usepackage{xcolor}
\usepackage{subfigure}
\usepackage{comment}
\usepackage[ruled,vlined]{algorithm2e}
\usepackage{todonotes} 
\usepackage{url}

\begin{document}

\title{A Novel Resampling Technique for Imbalanced Dataset Optimization}
%
%
\author{Ivan Letteri\inst{1}\orcidID{0000-0002-3843-386X} \and
Antonio Di Cecco\inst{2}\orcidID{0000-0002-9070-4663} \and
Abeer Dyoub\inst{3}\orcidID{0000-0003-0329-2419} \and
Giuseppe Della Penna\inst{4}\orcidID{0000-0003-2327-9393}}
\authorrunning{I. Letteri et al.}
%
\institute{Department of Information Engineering, Computer Science and Mathematics, University of L'Aquila, Italy \email{\{ivan.letteri,abeer.dyoub,giuseppe.dellapenna\}@univaq.it} \and
School Of AI, Italy \email{ant.dicecco@gmail.com} 
}
\maketitle              
\begin{abstract}
Despite the enormous amount of data, particular events of interest can still be quite rare \cite{Letteri2020Journal}. Classification of rare events is a common problem in many domains, such as fraudulent transactions, malware traffic analysis and network intrusion detection \cite{Letteri2018CSS}. For example, in the malware analysis domain, the amount of traffic generated from malware on the network is typically a very small fraction of the total traffic.

In a real network traffic scenario, where the number of flows generated by traffic malware is low compared to legitimate applications, we propose a prototype selection technique \cite{Letteri2019Caianiello} for dynamically imbalance dataset. Many studies have been developed for malware detection using machine learning approaches on various datasets, but as far as we know only the MTA-KDD'19 dataset has the peculiarity of updating the representative set of malicious traffic on a daily basis. This daily updating is the added value of the dataset, but it translates into a potential due to the class imbalance problem that the RRw-Optimized MTA-KDD'19 will occur. 

Although several methods for improving classifiers have been introduced, the identification of conditions for the efficient use of the particular method is still an open research problem.

In our paper, we capture difficulties of class distribution in real datasets by considering four types of minority class examples: safe, borderline, rare and outliers. In this work, we developed two versions of Generative Silhouette Resampling 1-Nearest Neighbour (G1Nos) oversampling algorithms for dealing with class imbalance problem. The first module of G1Nos algorithms performs a coefficient-based instance selection silhouette identifying the critical threshold of \textit{Imbalance Degree}. (ID), the second module generates synthetic samples using a SMOTE-like oversampling algorithm. The balancing of the classes is done by our G1Nos algorithms to re-establish the proportions between the two classes of the used dataset. The third module verifies our G1Nos algorithms by comparing it with two of the most used SMOTE-like oversampling algorithms evaluating the diagnostic ability of a binary classifier system (Multi Layer Perceptron).

The experimental results show that our oversampling algorithm work better than the other two SOTA methodologies in all the metrics considered.


\end{abstract}

\section{Introduction}


Unbalanced datasets, which lead to the ``curse of imbalanced train set'' \cite{Kubat97} is not a new problem. Class imbalances have been observed in many application problems such as detection of credit fraud, analysing financial risk, predicting technical equipment failures, managing network intrusion, text categorization and information filtering \cite{5128907}.  

Machine learning (as well as any statistical methodology) faces a formidable problem when dealing with imbalanced datasets. Indeed, classifiers are biased toward the majority classes and tend to misclassify minority class examples.

Therefore, typically the number of samples needs to be balanced before such methodologies can be successfully applied in imbalance datasets \cite{LOPEZ2013113}. 





Many techniques have been developed to counteract imbalanced data, both on the classifier level and on the data level. Algorithm level approaches adapt existing classifiers to bias the learning towards the minority class \cite{Lin2000Yoonkyung}\cite{Liu2000Bing}\cite{Pazzani1994} without changing training data. 
In particular, when dealing with classification models such as (deep) neural networks, usually the backpropagation algorithm is updated to weigh misclassification errors in proportion to the importance of the class (cost-sensitive neural networks) \cite{KukarK98} and/or the classification threshold is adjusted towards the minority class (\textit{threshold-moving}) \cite{1549828}. Both these techniques can be computationally expensive.

On the other hand, data level approaches, like the one proposed in this paper, change the class distribution by resampling the data space to improve the predictive performance.  In particular, we developed two algorithms, called G1No (\textit{Generative Resampling 1-Nearest Neighbour}) and G1No Gourmet, to compensate imbalance datasets by generating synthetic samples. The first applies a sort of greedy approach, the other choose samples in a more refined manner. 

To verify our approach, we use the MTA-KDD'19 dataset \cite{itasec2020}, which consists of two classes, where the malware traffic class is automatically daily updated. 

This uniqueness has made possible the study of an artificially induced ``dynamic unbalancing'' with respect to the balanced starting dataset.

The approach proposed in this paper is a data level one, thus, here and also for the lack of space, we concentrate only on these kind of approaches, which are classified into undersampling techniques and oversampling techniques. Both have advantages and disadvantages.

Reweigthing is typically used, but it can only be applied to base learning models that are designed to handle simple weights. On the other hand, resampling can be applied to any base learner \cite{4669722}, for this reason, we focused on it.

Random undersampling deletes examples from the majority class and can result in losing information invaluable to a model. A common problem in oversampling is the occurrence of oversampling due to the repeated addition of data, which makes the decision limit more stringent. Therefore, in its development, the oversampling method only copies the same data instead of generating new similar data.

Our method proposes to adjust the probabilities based on the difference in bad rates between the training and testing datasets. 

Data level approaches are based on the idea of resampling. The two main methods for resampling are undersampling and oversampling. Although many papers about the class imbalance problem have been written, most of them focus on either resampling techniques like Synthetic Minority Over-sampling (SMOTE) and ADAptive SYNthetic (ADASYN).

The entire experimentation process in this work consists of two methods combination: First one starts inducing unbalancing in the dataset to find the imbalance distribution of classes where the NN, used for testing, can be more resilient to the ``\textit{Curse of Imbalance Datasets in Machine Learning}'' \cite{LemaitreNA16}. 

We realized an \textit{Instance Selection} (IS) strategy to delimit the highest threshold at in order to have an imbalance dataset to use as benchmark for the experiments. This IS technique is based on silhouette coefficient (SC) and not classically based for an instance level analysis of data complexity \cite{Smith2014Martinez}, which requires expensive computation.

Second method applies an Oversampling technique for counterbalancing the imbalance dataset examined.

In our experiments, we use the optimised version of the MTAKDD'19 dataset \cite{itasec2020} composed by a restricted number of eleven features improved with a vector called RRw score \cite{letteri2020dataset}. As far as we know, it is the only dataset to have the fundamental peculiarity that made this study possible. MTA-KDD'19 consists of two classes (legitimate and malware), the malware traffic class is automatically daily updated. This uniqueness has made possible the study of an artificially induced ``dynamic unbalancing'' with respect to the balanced starting dataset.


Finally, we used a Neural Network, more precisely the usual MLP algorithm with the same architecture in \cite{itasec2020}, to predict the malware traffic with the new G1No's balanced datasets (generated), comparing the performance achieved respect to the same balancing distribution obtained with the two major oversampling algorithms discussed in the state of the art: SMOTE and ADASYN. 

The rest of this paper is structured as follows. Section \ref{sect:prelim} first summarized the preliminaries with related work. As the main contribution of this study, the Section \ref{sect:silhDDI} defines our methodology describing phases and relative steps of Undersampling Instance Selection Silhouette-based to induce artificial imbalance. In Section \ref{sect:2overDSs}, we expose out G1Nos algorithms to oversampling process on RRw-Optimized MTA-KDD experimental dataset described in Section \ref{sect:dataval}.

\section{Preliminaries}\label{sect:prelim}

In this section we recall several concepts and algorithms to give the reader some background abount the issue addressed by the present paper and support the development of our novel technique in the next sections.

\subsection{Dataset imbalance}

In general, dataset imbalance can be \textit{absolute} or \textit{relative} \cite{5128907}. In the first case, minority class samples are definitively scarce and underrepresented, whereas in the second minority samples are well represented but are outnumbered by majority class samples. The imbalance can be \textit{between-class}, when the number of samples representing a class differs from the number of samples representing the other class, or \textit{within-class}, when a class is composed of a number of different subclusters which, in turn, do not contain the same number of samples \cite{Japkowicz2001}.

Both these imbalance types are instances of the general problem known as the problem of \textit{small disjuncts} \cite{Japkowicz2003ClassIA}\cite{Ting1994ThePO}: since classification methods are typically biased towards classifying large disjuncts (that cover a large number of samples) accurately, they tend to misclassify the samples represented by small disjuncts.

In other words, a significant class imbalance can give rise to many problems when the dataset is used with learning algorithms \cite{Weiss2004}. Indeed, most classifiers work better on balanced datasets, and often this is assumed as a prerequisite for a correct learning \cite{Batista2004}. With an imbalanced dataset, the classifier is likely to be biased towards the majority classes: it has been shown that class imbalance seriously hinder the recognition of minority classes \cite{Napierala2010}, since the minority class samples may be insufficient to represent the boundaries between the two classes \cite{75512}. Moreover, imbalanced datasets are more deeply impacted by noisy data \cite{5680814}.
Therefore, data balancing techniques have been proposed to handle this scenario \cite{VanHulse2007}.

\subsection{Dataset (re)sampling}
\textit{Data sampling} consists of selecting some part of the population to observe so that, from the selected part, that we shall call the \textit{dataset}, one may estimate something about the whole population. There are many different types of data sampling methods that can be used, and there is no single best method to apply on all classification problems and with each classification model. 

Once built through appropriate sampling, a dataset may be \textit{resampled}  to improve it. In particular, resampling is commonly used to adjust the class distribution when dealing with unbalanced datasets.

\begin{figure}[!ht]
	\centering
	\includegraphics[width=0.7\hsize]{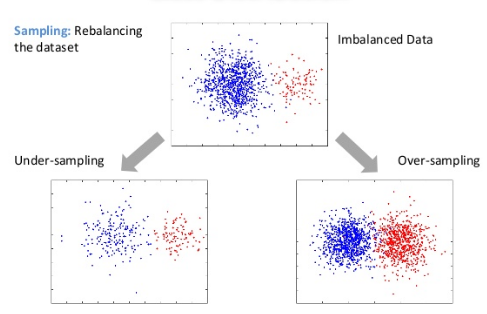}
	\caption{Dataset Resampling (source \cite{Arafat2018MachineLF}).}
	\label{fig:dr}
\end{figure}
There are two major techniques to achieve resampling, namely \textit{undersampling}, and \textit{oversampling}. Fig. \ref{fig:dr} shows graphically the effect of these techniques on an imbalanced dataset, which we will describe in detail in the following sections.

\subsubsection{Undersampling methods.}
Undersampling methods resample the data in order to reduce or delete some samples from the majority class. Clearly, when the number of samples in the minority class is too small compared to that of samples in the majority class, undersampling techniques are always ineffective.

Undersampling can be achieved by randomly removing samples (\textit{Random Undersampling} \cite{7300975})  or by applying statistical information (\textit{Informed Undersampling}) as with the Tomek Links \cite{4309523} or Neighborhood Cleaning Rule \cite{Laurikkala2001}.

The disadvantage of all the undersampling techniques above is that they could remove potentially useful data samples that could be important for the learning process. To cope with this issue, it is possible to use instance selection algorithms, where only samples considered less relevant are eliminated \cite{NietoYV20}. Generally speaking, an instance selection algorithm selects a subset of the samples that preserves the underlying distribution, so that the remaining data is still representative of the characteristics of the overall data. In particular, when dealing with data classification, the algorithm should identify a subset of the total available data to achieve the original purpose of the machine learning (or data mining) application, as if the whole data had been used \cite{Olvera2010Carrasco}. 
The \textit{optimal} outcome of an instance selection algorithm would be the minimum data subset that can accomplish the same task with no performance loss. Therefore, every instance selection strategy should deal with a trade-off between the reduction rate and classification quality. 

This technique is commonly applied also to reduce the original dataset to a manageable volume, to decrease the computational costs of the learning process \cite{Garcia2015Salvador}, and to remove noisy samples before applying learning algorithms in order to improve the classification accuracy.

\subsubsection{Oversampling methods.} \label{subsec:ovsampling}
Oversampling methods add samples to the minority classes in order to make their size similar to the one of the majority class.
Such new samples can be original data, randomly chosen from the data source, or synthetic data, generated randomly (e.g., by duplicating existing samples) or with specific algorithms such as SMOTE  \cite{Chawla2002} or ADASYN \cite{4633969}.

\begin{figure}[!ht]
	\centering
	\includegraphics[width=0.6\hsize]{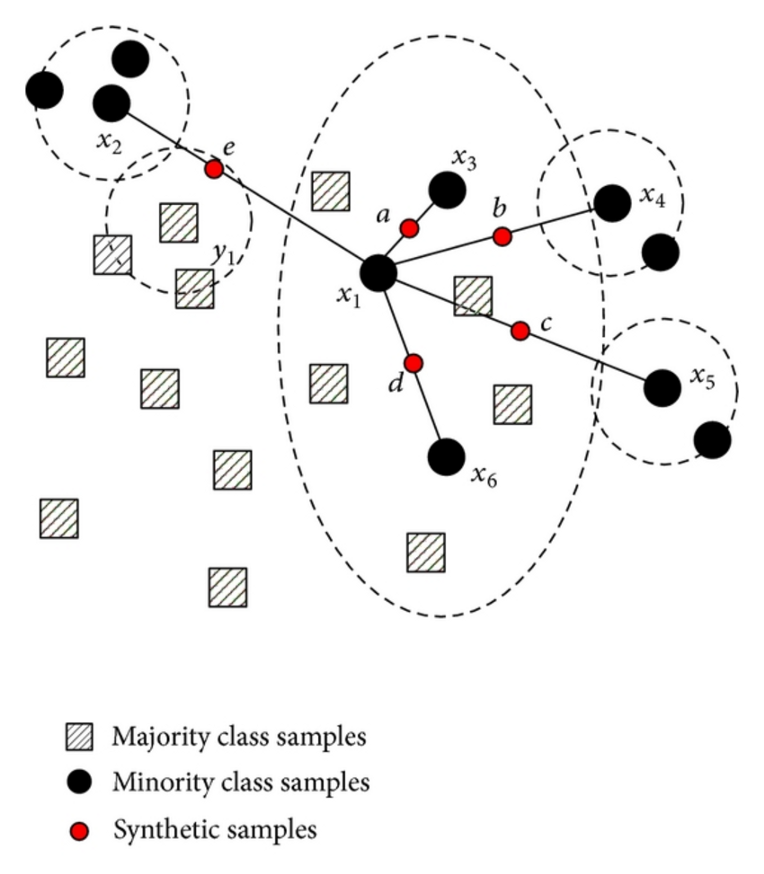}
	\caption{Synthetic Instance generation by SMOTE algorithm [source \cite{Hu2013Feng}].}
	\label{fig:synthInstances}
\end{figure}

In particular, SMOTE tries to avoid the risk of overfitting, commonly faced by random oversampling, by exploiting the nearest neighbours, as shown in Figure \ref{fig:synthInstances}. First, it chooses a random minority observation $x_i$, and then it randomly selects an instance $x_u$ among the $k$ nearest minority class neighbours of $x_i$. Finally, a new random sample $s_i$ is generated by interpolating the two samples: $s_i = x_i + w \times (x_u - x_i)$, where $w$ is a random weight in $[0,1]$.


The oversampling strategy of SMOTE allows to effectively dwell with between-class imbalance, but ignores within-class imbalance.  Moreover, the algorithm does not specifically enforce the decision boundary: instances far from the class border are oversampled with the same probability as those close to the boundary, whereas it has been argued that classifiers could benefit from the generation of samples closer to the class border \cite{Douzas2017Geo}.

To this aim, the ADASYN algorithm \textit{adaptively} generates minority data samples privileging the samples which are near the class boundaries. In ADASYN, the sample generation technique is similar to SMOTE, but this time \textit{for each} minority class data example $x_i$, $g_i= r_i \times g$ interpolated samples are generated (whereas SMOTE generates the same number of samples for some randomly-chosen minority samples), where $g$ is the total number of synthetic data samples needed and $r_i$ is proportional to the number of majority class samples between the $k$ nearest neighbours of $x_i$ (so it is highest near the class boundaries). 
    
\begin{figure*}[!ht] 
    \centering
    \subfigure[ADASYN]{%
        \includegraphics[width=0.45\hsize]{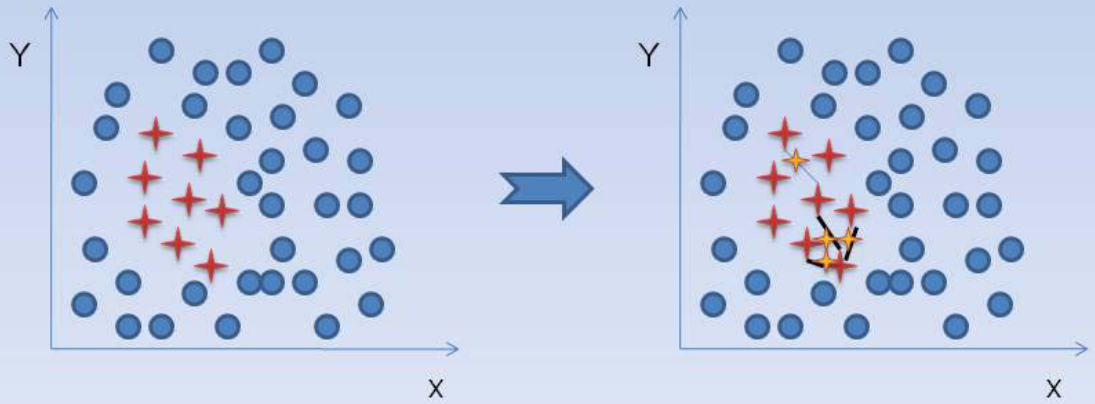}%
        \label{fig:algoAdasyn}}%
    \quad%
    \subfigure[SMOTE]{%
        \includegraphics[width=0.45\hsize]{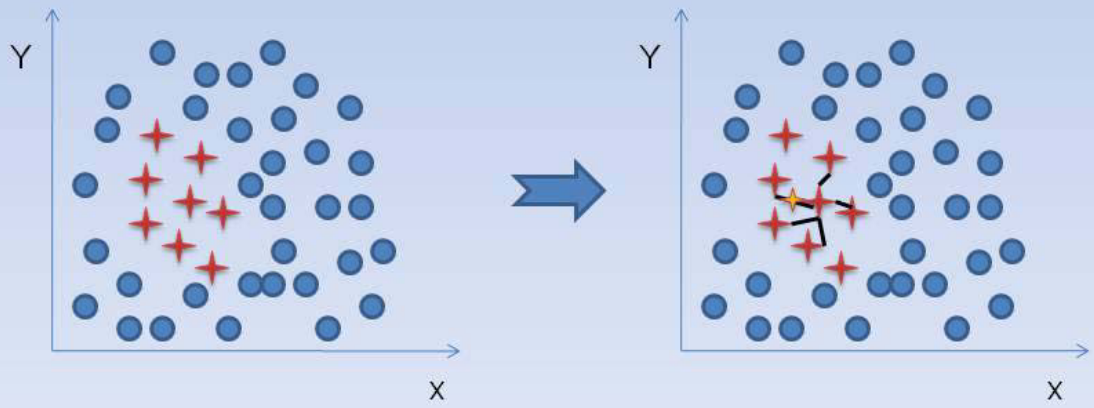}%
        \label{fig:algoSmote}}%
    \caption{Difference between ADASYN and SMOTE algorithms.}
    \label{fig:compalgos}
\end{figure*}

Fig. \ref{fig:compalgos} shows the key difference between ADASYN (\ref{fig:algoAdasyn}) and SMOTE (\ref{fig:algoSmote}).

The main disadvantage of such oversampling techniques is that they assume that the space between any two minority class samples belongs to the minority class, which may not be always true when the data is not linearly separable.


A common problem in oversampling is the occurrence of oversampling due to the repeated addition of data, which makes the decision limit more stringent. Therefore, in its development, the oversampling method only copies the same data instead of generating new similar data.

Moreover, it is important to note that ovesampling cannot be directly applied on the entire data set, for this reason we split in Train, Test, and Validation the dataset appling the oversampling only to the Train set because, if the data is first oversampled and then split into validation and train subsets, there is a high chance for the same data being present in test as well as training set, biasing the decision model evaluation.

\subsection{K Nearest Neighbors (KNN)}
\begin{figure}[!ht]
	\centering
	\includegraphics[width=0.7\hsize]{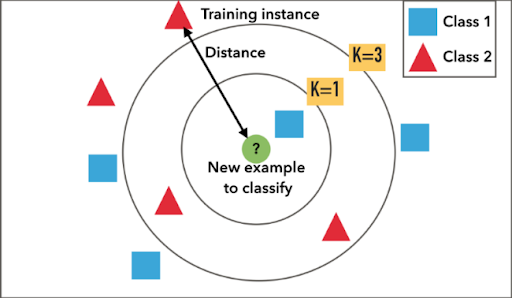}
	\caption{k-Nearest-Neighbours (source \cite{knn1}).}
	\label{fig:knn}
\end{figure}

K Nearest Neighbors is a very simple classification algorithm, where a sample is assigned to the class that is most frequent among the $k$ training (i.e., already correctly classified) samples nearest to it. In a binary classification problem, as normal/malicious traffic in our case, an odd $k$ is useful to avoid ties (see fig. \ref{fig:knn}).

In particular, later in this paper we will make use a basic version of this algorithm, where $k=1$. That is, we assign to a new sample the same classification of its nearest training sample. Even if it may appear simplistic, this approach is useful when dealing with imbalanced training sets, where a class has a more dense distribution of samples. Indeed, in this case, there is an high probability that \textit{any} sample has \textit{many} training samples of the majority class around it, so it will get \textit{always} classified in such a class. On the other hand, by restricting to a single nearest neighbour, we try to find the class of the sample that is \textit{the most similar} to the new one. In other words, we try to have the \textit{maximum possible sensibility towards the minority class}.

\subsection{Silhouette Coefficient} \label{sec:silhouette}
Let $C$ be dataset that has been partitioned, via supervised learning classification, into two clusters $C_1, C_2$. Then, for every sample $t \in C$, with $t \in C_i$, we calculate the mean distance between $t$ and all other samples in the same cluster as follows:

$$
a(t) = \frac{1}{|C_i| - 1} \sum_{t' \in C_i, t' \neq = t} d(t,t')  
$$

\noindent where $d(t,t')$ is the (Euclidean) distance between the samples $t$ and $t'$ in the cluster $C_i$.

We can interpret $a(t)$ as a measure of how well $t$ is assigned to its cluster (the smaller the value, the better the assignment). Indeed, $a(i)$ can be also called \textit{mean dissimilarity} between $t$ and all the other samples in $C_i$. Similarly, we can calculate the mean dissimilarity between $t$ and all the samples in another cluster $C_k$, and extract the smallest mean dissimilarity with respect to \textit{any other cluster} $C_k \neq C_i$ as follows:

$$
b(t) = \min_{k \not = i} \frac{1}{|C_k|} \sum_{t' \in C_k} d(t,t')
$$

The cluster with this smallest value is said to be the \textit{neighboring cluster} of $t$ because it is the next best fit cluster for $t$. 


The \textit{silhouette coefficient} of $t$ is then defined as

$$
s(t) = \left\{ \begin{array}{cc}
     \frac{b(t) - a(t)}{\max\{a(t) , b(t)\}} & |C_i| > 1 \\
     0 & |C_i|=1
\end{array}\right.
$$

Roughly speaking, $s(t)$ measures how similar (close) $t$ is to its own cluster (i.e., samples in the same cluster) compared to the other clusters. More in detail, $-1 \leq s(t) \leq +1$ and a value near $1$ indicates that $t$ is far away from the neighbouring clusters (so its classification appears \textit{good}), a value of $0$ indicates that the sample is on or very close to the \textit{decision boundary} between two neighbour clusters, and negative values indicate that the sample might have been assigned to the wrong cluster. Similarly, the mean silhouette of all the samples in a particular cluster can be used to assess the overall quality of such a cluster.


It is worth noting that the definition of silhouette is purely metric (like for 1NN) and independent of any model: 
however, the silhouette coefficient offers an effective way of enhancing the performance of any classification model by simply giving more weight  (\textit{sensibility}) to the samples with smaller silhouette (\textit{near the decision boundary}). Indeed, such samples mark the class boundaries better than all the other samples, so they should not be eliminated, but rather the model should learn to use them. 




\section{Silhouette-driven Dataset Imbalancing}\label{sect:silhDDI}

In general, in any updated malware traffic dataset the number of malware samples tends to grow faster than the legitimate one. In particular, our reference dataset, the MTA-KDD'19 \cite{MTAKDD19}, is currently balanced, but the authors expect it to become (heavily) unbalanced in the near future since it is constantly fed with new malware traces.


Therefore, to obtain an effective dataset to be used in our re-balancing experiments, we anticipated this phenomenon by manually introducing an imbalance in the  MTA-KDD'19 dataset. However, we want a valid, realistic and non-biased result dataset, thus we did not choose to oversample the malware traffic class (introducing synthetic samples that may not be completely realistic), but we undersampled the legitimate traffic class. Moreover, we did not perform a simple random undersampling, but we applied a silhouette-driven undersampling, to avoid removing highly informative samples from the dataset, as discussed in Section \ref{sec:silhouette}. 



Let $D=\{(x_1,y_1),\ldots,(x_m,y_m)\}$ be a (training) dataset where, in each item $(x_i,y_i)$, $x_i$ is the sample (vector in the feature space) and $y_i$ its assigned class (label). Let $m_{min}$ and $m_{max}$ be the number of minority and majority class samples, respectively, such that $m_{min} \leq m_{max}$ and $m_{min} + m_{max} = m$.\\

We call \textit{Imbalance Degree} (ID) of $D$ the ratio $d = \frac{m_{min}}{m_{max}}$. Then, given a classification model, we call Imbalance Degree fall-down threshold (IDft) of such a model the maximum ID  which the performance of the classification model falls under an acceptable level (which depends, of course, from the specific classification experiment). Roughly speaking, the IDft measures the robustness of the model with respect to the class imbalance on the training dataset.

\begin{figure}[!ht]
	\centering
	\includegraphics[width=0.7\hsize]{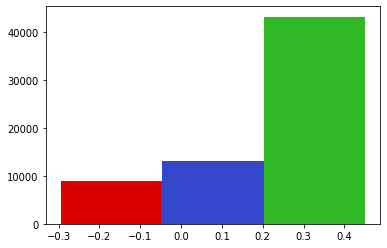}
	\caption{Silhouette coefficient distribution over samples}
	\label{fig:silhDistrib}
\end{figure}

Fig. \ref{fig:silhDistrib} shows the silhouette coefficient distribution over the samples in our original dataset. The red bar counts the samples with a silhouette near $-1$ (13.75\%), the blue bar the ones  close to $0$ (20.12\%), and the green one the samples with the highest silhouette (66.12\%).

Analysing in-depth the samples belonging to the "red class", we note that they mostly (96.52\%) belong to the \textit{legitimate traffic} class. This may suggest to further investigate the quality of the legitimate traffic source that is used in our dataset (i.e., the pcap files marked as \textit{Normal} in the Malware Capture Facility Project (MCFP) belonging to the Stratosphere project \cite{MCFP}). Anyway, if we suppose this classification to be valid, we may say that the samples with low silhouette are somehow \textit{atypical} for the cluster they are in, and therefore they convey a lot of information (are hard to classify), i.e., removing them from the dataset  may drastically decrease the classification performance of any model.

To further verify this assumption, we ordered the samples in the minority class of the dataset (legitimate traffic samples) with respect to their silhouette coefficient. Then, we progressively removed the from 5\% to 95\% of such legitimate samples, in descending silhouette order. The resulting training set is shuffled, and the removed samples are moved to the validation set.

After each iteration, we measured the decrease of the classification performance using the same MLP described in \cite{itasec2020} (a rectangle-shaped fully connected MLP with two hidden layers, both with $22$ neurons, and a single-neuron output layer with sigmoid activation function, trained for 10 epochs with batch size equal to 10), whose execution on the full dataset will be our baseline.

To make our experimentation more robust, we replicated the same experiment using a 5-cross fold validation of the dataset and we calculated the average of the metrics. \\

\begin{table}[!ht]
\centering
\caption{Progressive imbalance of MTA-KDD'19 dataset removing legitimate samples with higher silhouette coefficient.}
\begin{tabular}{|c|c|c|c|c|c|c|}
\hline
\textbf{Iteration} & \textbf{\% Malware} & \textbf{\% Legitimate} & \textbf{Precision} &  \textbf{F-measure} & \textbf{Accuracy} & \textbf{AUC} \\ \hline
\textit{1st}   &   55.06               & 44.94                  & 99.5782   & 99.5175     & 99.6887           & 99.9803      \\ \hline
\textit{2nd}   &   56.40               & 43.60                  & 99.3081   & 99.5497     & 99.5709           & 99.9838      \\ \hline
\textit{3rd}   &   57.80               & 42.20                  & 99.6895   & 99.6994     & 99.7066           & 99.9805      \\ \hline
\textit{4th}   &   59.27               & 40.73                  & 99.8212   & 99.6439     & 99.8307           & 99.9908      \\ \hline
\textit{5th}   &   60.82               & 39.18                  & 99.6865   & 99.5841     & 99.7718           & 99.9813      \\ \hline
\textit{6th}   &   62.45               & 37.55                  & 99.8225   & 99.6547     & 99.7718           & 99.9858      \\ \hline
\textit{7th}   &   64.17               & 35.83                  & 99.6429   & 99.6729     & 99.8118           & 99.9945      \\ \hline
\textit{8th}   &   65.99               & 34.01                  & 99.6653   & 99.6447     & 99.8037           & 99.9977      \\ \hline
\textit{9th}   &   67.911              & 32.09                  & 99.6885   & 99.2455     & 99.8215           & 99.9912      \\ \hline
\textit{10th}   &   69.96               & 30.04                  & 99.5946  & 99.6078     & 99.8324           & 99.9959      \\ \hline
\textit{11st}   &   72.12               & 27.88                  & 99.6922  & 99.6676     & 99.8377           & 99.9934      \\ \hline
\hline \hline
\textit{\textbf{12nd}}   &   \texttt{74.42}               & \texttt{25.58}     & \textit{95.6465}    & 33.8705        & \textit{99.0711}           & 99.4714      \\ \hline \hline 
\textit{13rd}   &   \texttt{76.88}               & \texttt{23.12}      & \textbf{21.4012}   & 33.7381        & \textbf{35.3404}           & 95.3507      \\ \hline \hline \hline
\textit{14th}   &   79.51               & 20.49                  & 19.9795     & 33.6861       & 32.8572           & \texttt{45.7322}     \\ \hline
\textit{15th}   &   82.32               & 17.68                  & 17.5616     & 29.9482     & 25.1221           & 42.9722      \\ \hline
\textit{16th}   &   85.34               & 14.66                  & 16.8386     & 28.7407     & 24.7668           & 39.3858      \\ \hline
\textit{17th}   &   88.59               & 11.41                  & 16.2244     & 22.2169     & 24.5455           & 53.1283      \\ \hline
\textit{18th}   &   92.09               & 7.91                   & 14.3026     & 29.4446     & 16.1845           & 46.8645      \\ \hline
\textit{19th}   &   95.88               & 4.12                   & 14.4489     & 28.2249     & 20.4009           & 52.6848      \\ \hline
\end{tabular}
\label{tab:MLPmetrics}
\end{table}

In Tab. \ref{tab:MLPmetrics}, we report the measured performance metrics, i.e., the \textit{Precision}, \textit{F-measure}, the \textit{Accuracy}, and the Area Under ROC Curve (AUC). Precision is more focused in the majority class than in the minority class, it actually measures the probability of correct detection of positive values. The \textit{F-Measure} provides a single score that balances both the concerns of precision respect to the true positive rate in one number. The Receiver Operating Characteristic (ROC) curve measures the ability to distinguish between the classes, and consists of the (TPR) and (FPR) through various probability thresholds where the \textit{True Positive Rate} (TPR)/\textit{True Negative Rate} (TNR) measure the proportion of  positives/negatives that are correctly/wrongly identified, respectively. The Area Under the ROC curve (AUC) is a measure of how well a parameter can distinguish between two classes (malware/legitimate).



\begin{figure*}[!ht] 
    \centering
    \subfigure[Silhouette Coefficient higher]{%
        \includegraphics[width=0.45\hsize]{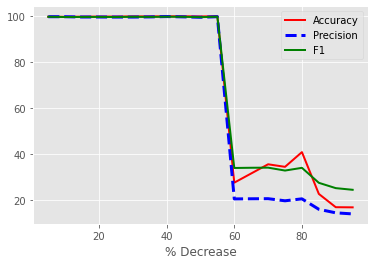}%
        \label{fig:RandFeat1}}%
    \quad%
    \subfigure[Silhouette Coefficient lower]{%
        \includegraphics[width=0.45\hsize]{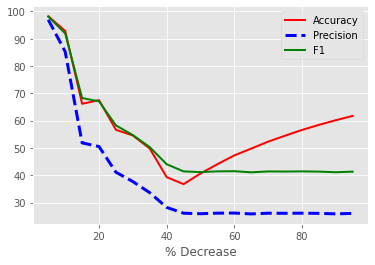}%
        \label{fig:RandFeat3}}%
    \caption{MLP metrics comparison of two opposite progressive decrease of legitimate class samples approaches.}
    \label{fig:RandFeat}
\end{figure*}


More in detail, Fig. \ref{fig:RandFeat1} shows that when 55\% of the legitimate samples with the highest silhouette are removed, reaching an ID of 0.3721 (72.88\% of Malware samples and  27.12\% of Legitimate samples), all the metrics fall down very quickly. On the other hand, if we remove the legitimate samples in reverse order, i.e., starting from the lowest silhouette, Fig. \ref{fig:RandFeat3} clearly shows that this dramatic decrease happens earlier, after a removal of about 10\% of the samples. 

\begin{figure}[!ht]
	\centering
	\includegraphics[width=0.9\hsize]{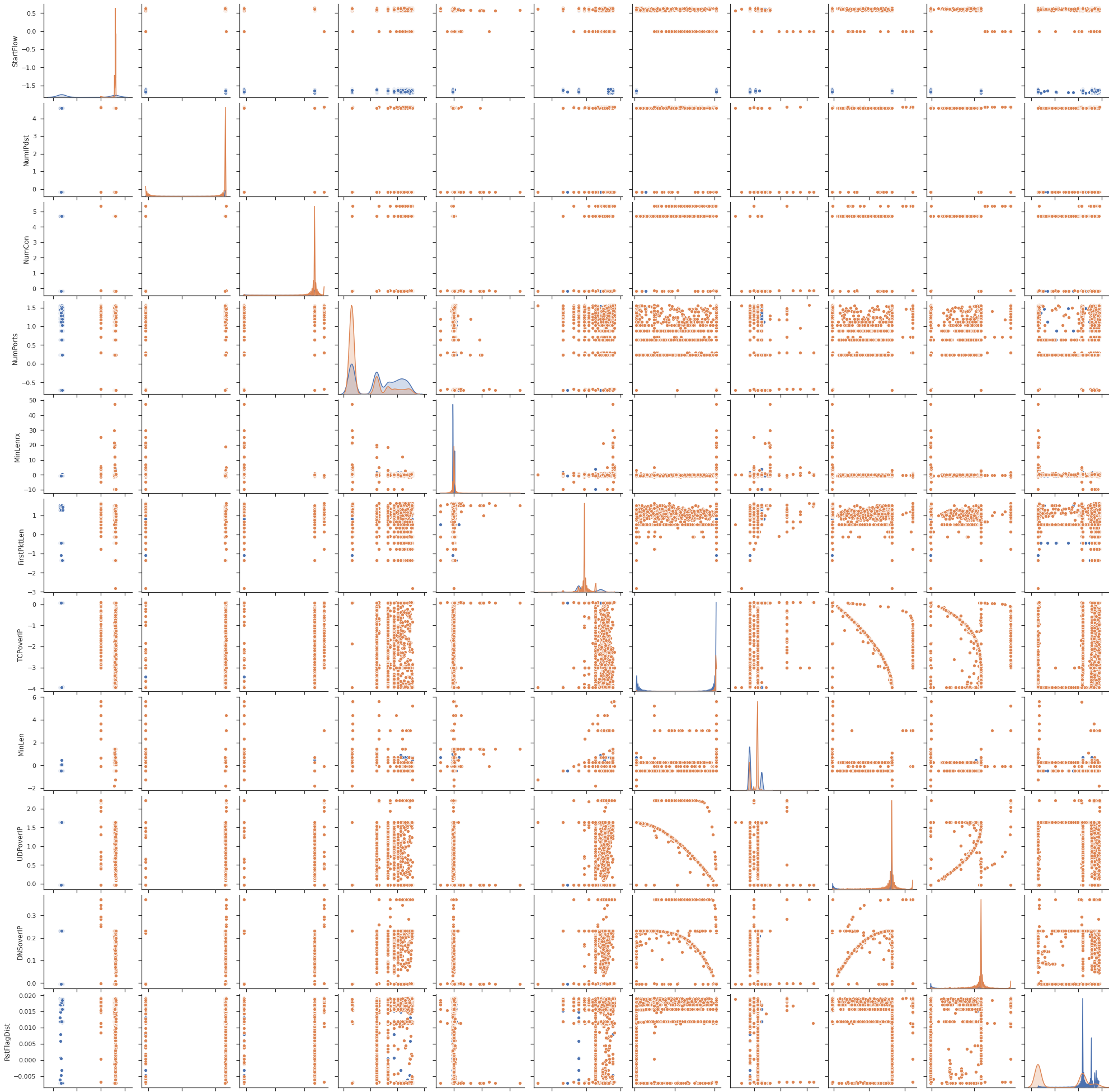}
	\caption{Pair Plot of unbalanced MTA-KDD'19 dataset.}
	\label{fig:rrwOptMTAKDD}
\end{figure}

Given these values, we decided to set the IDFt of our MLP to the one corresponding to the $12nd$ interaction, where the F-measure fall down drastically and the Precision decrease under 99\%, i.e., 0.3721. Therefore, our resulting unbalanced MTA-KDD dataset will be the one produced in such an iteration. 
The dataset can be graphically represented by the pair plot of Fig. \ref{fig:rrwOptMTAKDD}. 
Pair Plots show the relation between the dataset features (i.e., \textit{NumIPdst}, \textit{NumCon}, \textit{NumPorts}, \textit{MinLenrx}, \textit{1stPktLen}, \textit{MinLen}, \textit{UDPoverIP}, \textit{TCPoverIP}, \textit{DNSoverIP}, \textit{RstFlagDist} and \textit{FinFlagDist}: see \cite{itasec2020} for full details) and allow to see on the diagonals the density distributions of the samples. In the plots, blue dots represent legitimate traffic samples whereas red dots represent malware samples.

\section{G1No algorithms: two Novel Oversampling methods for balancing datasets}\label{sect:2overDSs}

In this section we present two novel oversampling methods for balancing datasets, namely G1No and G1No Gourmet. Such algorithms use a novel generative technique that is completely different, e.g., from the well known GANS \cite{Kiyoiti2019} or  or Variational Autoencoders \cite{Kingma2019}. 

In particular, our algorithms perform sampling directly on the imbalanced dataset using the shortest possible distance, i.e., the \textit{nearest neighbour}. 

Reducing the (Euclidean) distance neighbors closer and closer to the training data induce low bias, but the model will be much more dependent on the particular training examples chosen as consequence a high variance. In general, our hypothesis is that the 1NN choice will be the best also in other dataset types and we will investigate this choice. In fact our assumption is of bayesian type. In G1Nos algorithms, we generate samples with the prior that the underling process that gave the samples can be described with a 1NN model and we extract new samples via a kind of von Neumann rejection sampling. As stated the choice of 1NN is to have the maximum possible sensibility toward the minority class. Obviously this choice is also most sensible to noise but It's not the case in our work

There is one logical assumption here by the way, and that is our training set will not include same training samples belonging to different classes, i.e. conflicting information.

In 1NN the bias is low, because this model is fit only to the 1-nearest point. This means it is really close to the training data. To the other side, the variance is high, because optimizing on only 1-nearest point means that the probability that the noise in the data is really high. Since the test sample is in the training dataset, so it choose itself as the closest and never make mistake. 

The proposed methodology can be divided in three steps. First we train a 1 Nearest Neighbour (1NN) with the imbalanced dataset by splitting it in two segments, i.e., 25\% of the dataset as the validation set and the remaining 75\% as the training set.


Then we generate a set of random Gaussian samples to compensate the imbalance.
To this aim, given a dataset $D=\left\{s_1,\ldots,s_m\right\}$ with $n$ features $\left\{f_1,\ldots,f_n\right\}$ and $m$ samples, we make use of a Gaussian Random Number Generator (GRNG) function which follows the mean $\mu_{f_j}$ and a standard deviation $\sigma_{f_j}$ values of each feature $f_j$ derived from the dataset as follows.

Given a dataset $D$ with $n$ features and $m$ samples for every sample $s_i$ the G1No's processes applied to the two G1No algorithms follow steps:\\

\textbf{G1No's process}
\begin{itemize}
    \item \textit{Step 1} calculates the \textbf{Mean} for each $n$ features as follow: $\mu_{f_n} = {\frac {1}{m}}\sum _{i=1}^{m}s_{i}$.
    
    \item \textit{Step 2} calculates the \textbf{Standard Deviation} for each $n$ features as follow: $\sigma_{f_n} = \sqrt{\sum (f_n - \mu_{f_n})^2}$.
\end{itemize}

\textbf{G1No Gourmet's process}
\begin{itemize}
    \item \textit{Step 1} calculates the weight for overall samples as follow: $w_i = \frac{(silh_{max} - silh_i)}{(silh_{max} - silh_{min})}$. 
 
    \item \textit{Step 2} calculates the \textbf{Mean SC weighted} for each $n$ features as follow: $\mu_{f_n} = \frac{\sum (f_n * w_i)}{\sum w_i}$.
    
    \item \textit{Step 3} calculates the \textbf{Standard Deviation SC weighted} for each $n$ features as follow: $\sigma_{f_n} = \sqrt{\frac{\sum ((f_n - \mu_i) * w_i)^2}{\sum w_i}}$.
\end{itemize}

where $f_n$ is the feature $n$ and $w_i$ is the SC weight for every sample $i$. With $silh_i$ we denote the SC per sample $i$, $silh_{max}$ and $silh_{min}$ are the maximum and minimum SCs respectively.

After pre-processing of the two different Means and Variances, these two tuples are sent to GRNG algorithm that generates the random Gaussian samples to balancing the RRw-Optimized MTA-KDD dataset. Before adding a new sample to the dataset, we check it through the previously trained 1NN. If the 1NN recognises the sample as legitimate, it is added to the legitimate class, otherwise it is rejected.

It is worth noting that we chose a Gaussian distribution because we want to regularise the \textit{in-class} distribution of the samples. Indeed, it is well known that a Gaussian distribution of the samples produces simpler class boundaries and simpler models. So the choice of a GRNG impose an implicit restraint on the complexity of the models without an explicit use of regularizing  terms in the loss function. 

Finally, duplicate samples (i.e., already present in the dataset) are discarded, and the other ones are checked through the previously trained 1NN. If the 1NN recognises the sample as legitimate, it is added to the legitimate class, otherwise it is discarded. This process is repeated until we reach the desired class balance.

\begin{figure*}[!ht] 
    \centering
    \subfigure[Pair Plot of G1No]{%
        \includegraphics[width=0.48\hsize]{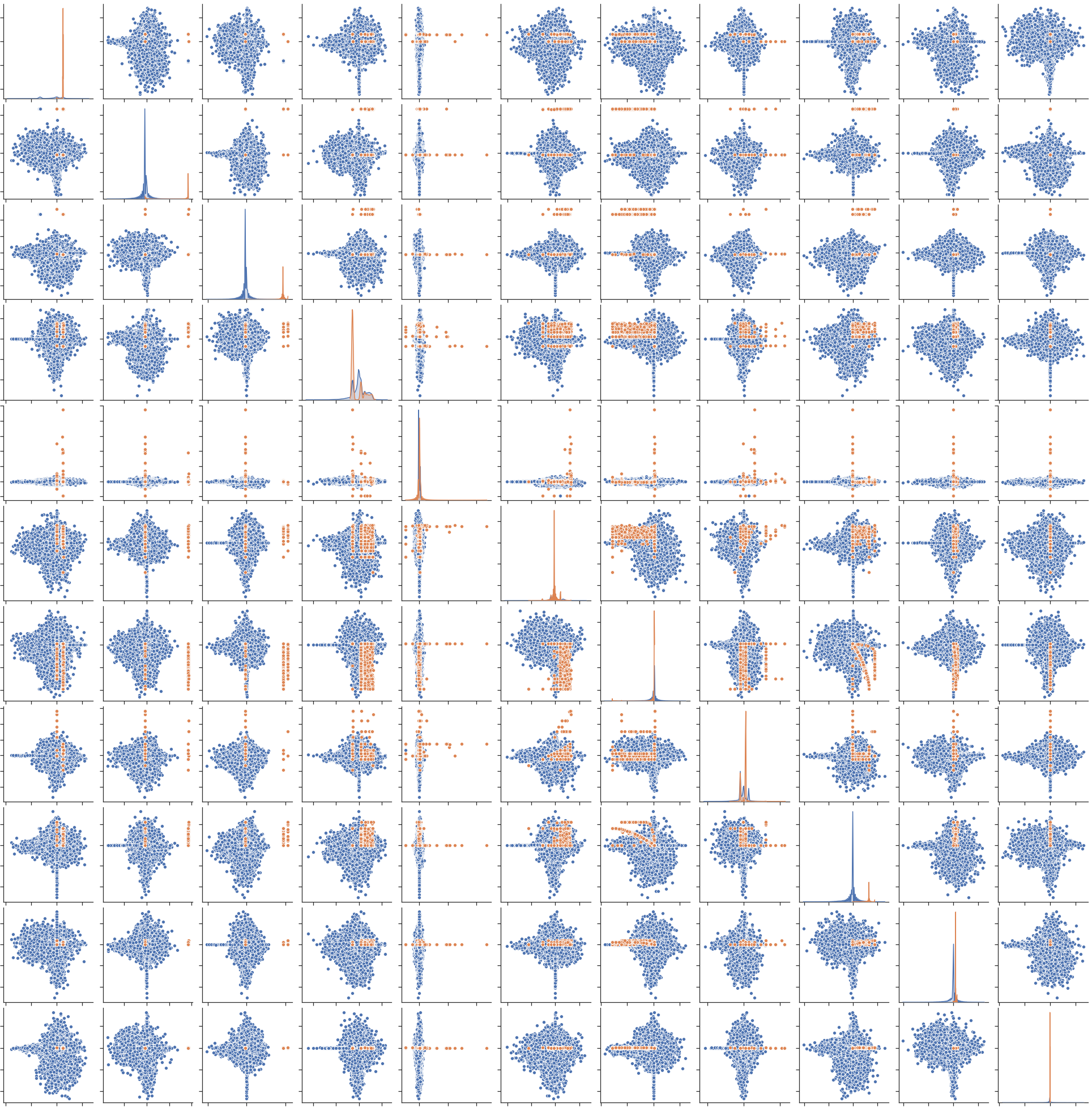}%
        \label{fig:PPG1No}}%
    \quad%
    \subfigure[Pair Plot of G1No Gourmet]{%
        \includegraphics[width=0.48\hsize]{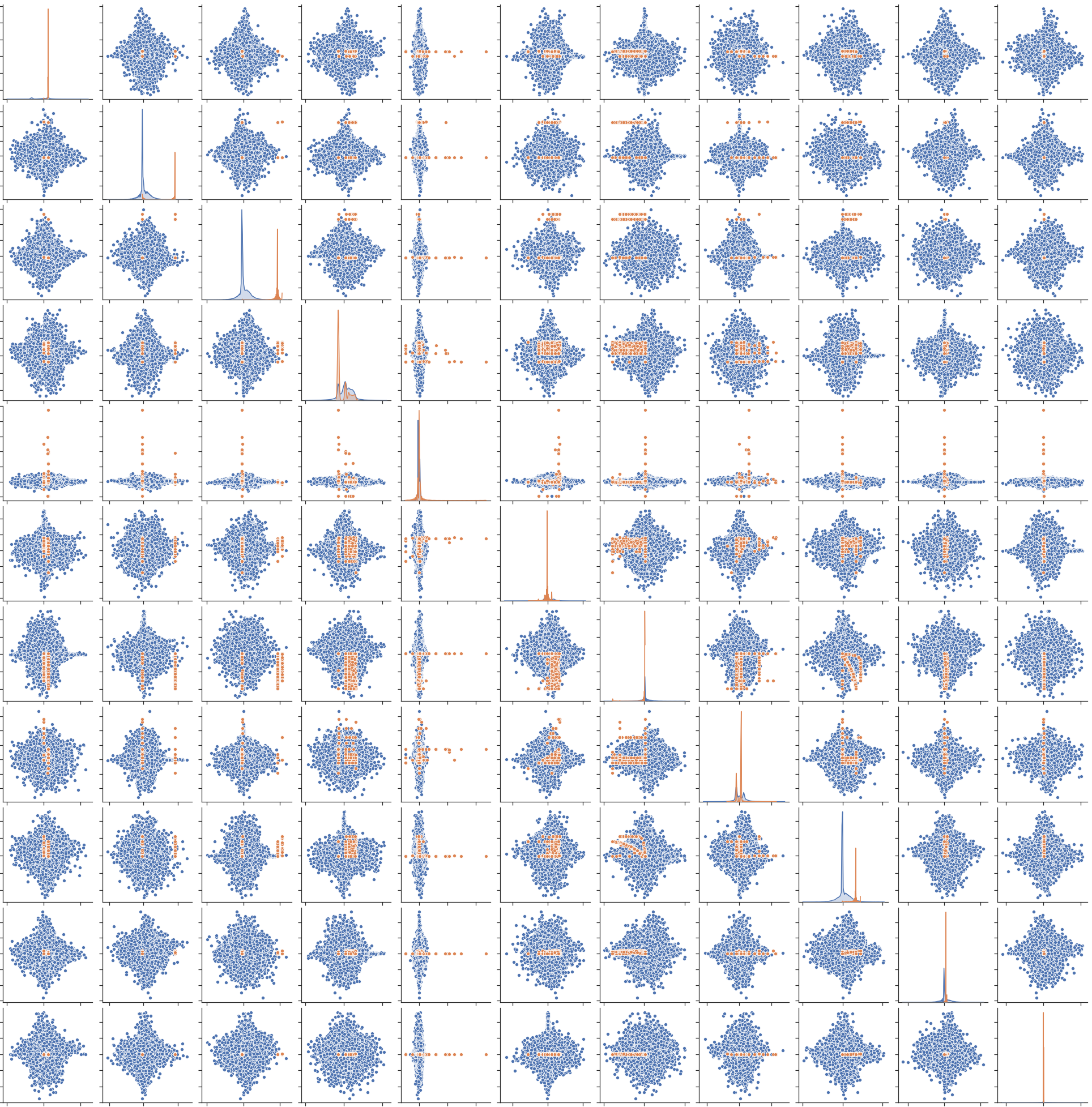}%
        \label{fig:PPG1Nog}}%
    \caption{Pair plot of the dataset re-balanced with G1No and G1No Gourmet algorithms.}
    \label{fig:PPG1Nos}
\end{figure*}

Fig. \ref{fig:PPG1Nos} shows the pair plots of the MTD-KDD unbalanced dataset shown in fig. \ref{fig:rrwOptMTAKDD}, re-balanced with G1No and G1No Gourmet, respectively.\\

\newpage
\subsubsection{G1No's algorithm}
\begin{itemize}
    \item a) Calculate the number of synthetic samples to be generated to compensate the minority class. So, given a set $G \longleftarrow \emptyset$ to fill with the random Gaussian synthetic samples in order to achieve the quantity $m = (m_{maj} - m_{min})$.
    \item b) Generate all the $g_i$ synthetic data examples according to the following steps:\\
        Do the Loop from $1$ to $m$:
        \begin{itemize}
        \item (i) \textit{GRGN G1No's} module generate the synthetic data example:\\ 
                $g_i = randomGauss(\mu(f_n),\sigma(f_n))$ for each $n$ features in the dataset $D$.
        \item (ii) Check $g_i$ with 1NN, if $g_i$ is classified as \textit{Legitimate} go to the next step, otherwise discard $g_i$ and go to step (i).
        \item (iii) if $g_i \notin G$, $G \longleftarrow g_i$ otherwise discard $g_i$.
        \end{itemize}
    \item c) return the set $G$ of $m$ Gaussian synthetic samples to concatenate to the minority class.
\end{itemize}

\begin{algorithm}[H]
\SetAlgoLined
\KwData{$GRNG = (m, n, \mu,\sigma)$}
\KwIn{Num. of samples $m$, Num. of feature $n$, Mean $\mu$, StdDev $\sigma$}
\KwOut{array G of synthetic samples}
$G = 0 , N = n$\
\BlankLine
\While{$m \neq 0$}{
  \BlankLine
  \For{$n\in N$}{
    $g_i \leftarrow$ random\_gaussian($\mu_{f_n}$, $\sigma_{f_n}$)
    \BlankLine
    \If{$g \notin G$}{ $G \longleftarrow g_i$\ }
   }
 $m = m-1$
 }
\caption{GRNG G1Nos module}
\label{algo:1NN}
\end{algorithm}


\section{Dataset validation}\label{sect:dataval}


In this section, we test the quality of the re-balanced MTD-KDD datasets obtained with our G1No algorithms (described in section \ref{sec:2overDSs}), comparing them with the datasets generated by the two state-of-the-art oversampling algorithms SMOTE and ADASYN introduced in Section \ref{subsec:ovsampling}.

The number of legitimate synthetic samples generated by each algorithm is 21541, so all re-balanced datasets contain a total amount of 70226 samples. 

\subsection{Pair plot comparison}

\begin{figure*}[!ht] 
    \centering
    \subfigure[Pair Plot of SMOTE]{%
        \includegraphics[width=0.45\hsize]{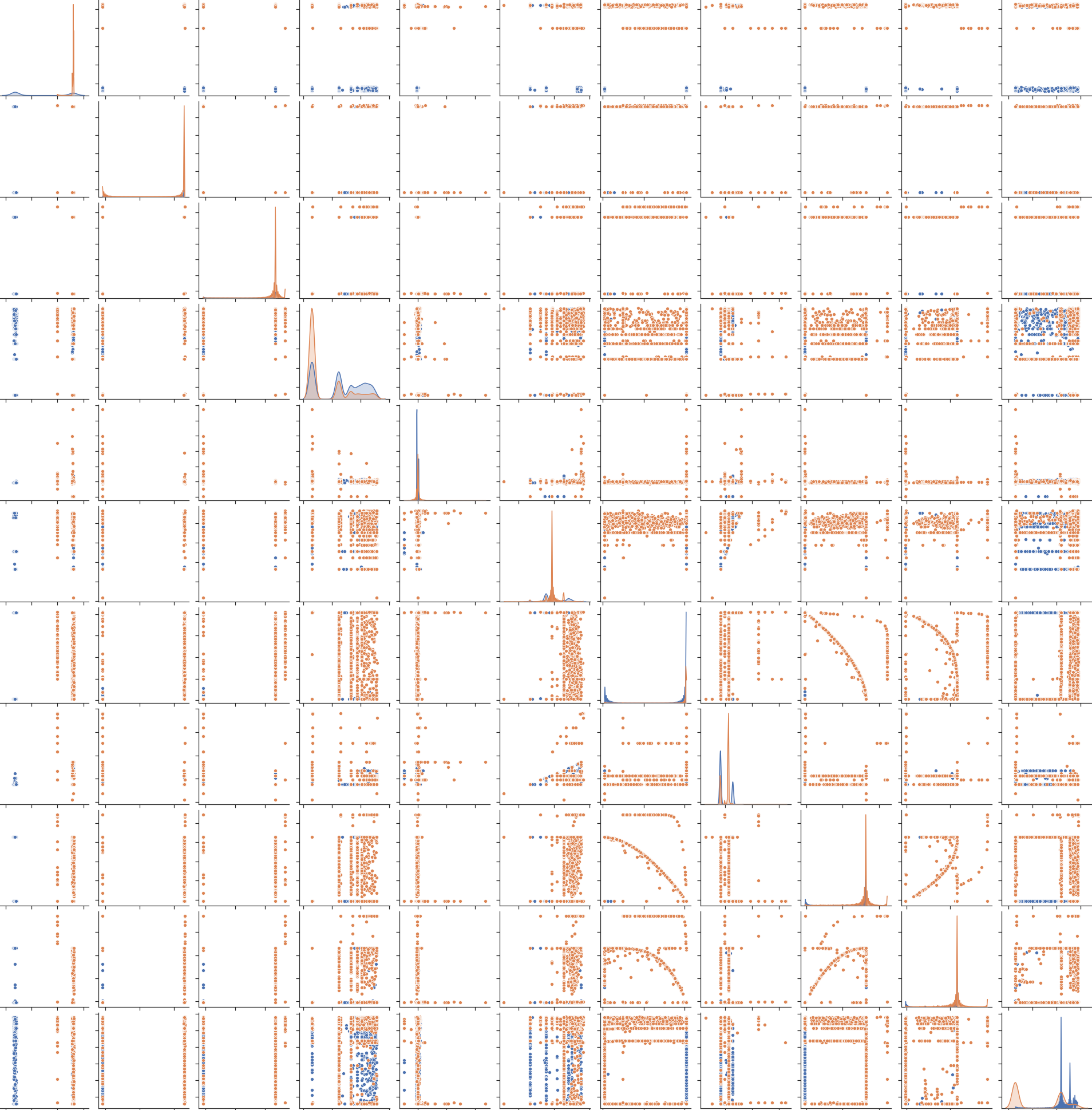}%
        \label{fig:PPSMOTE}}%
    \quad%
    \subfigure[Pair Plot of ADASYN]{%
        \includegraphics[width=0.45\hsize]{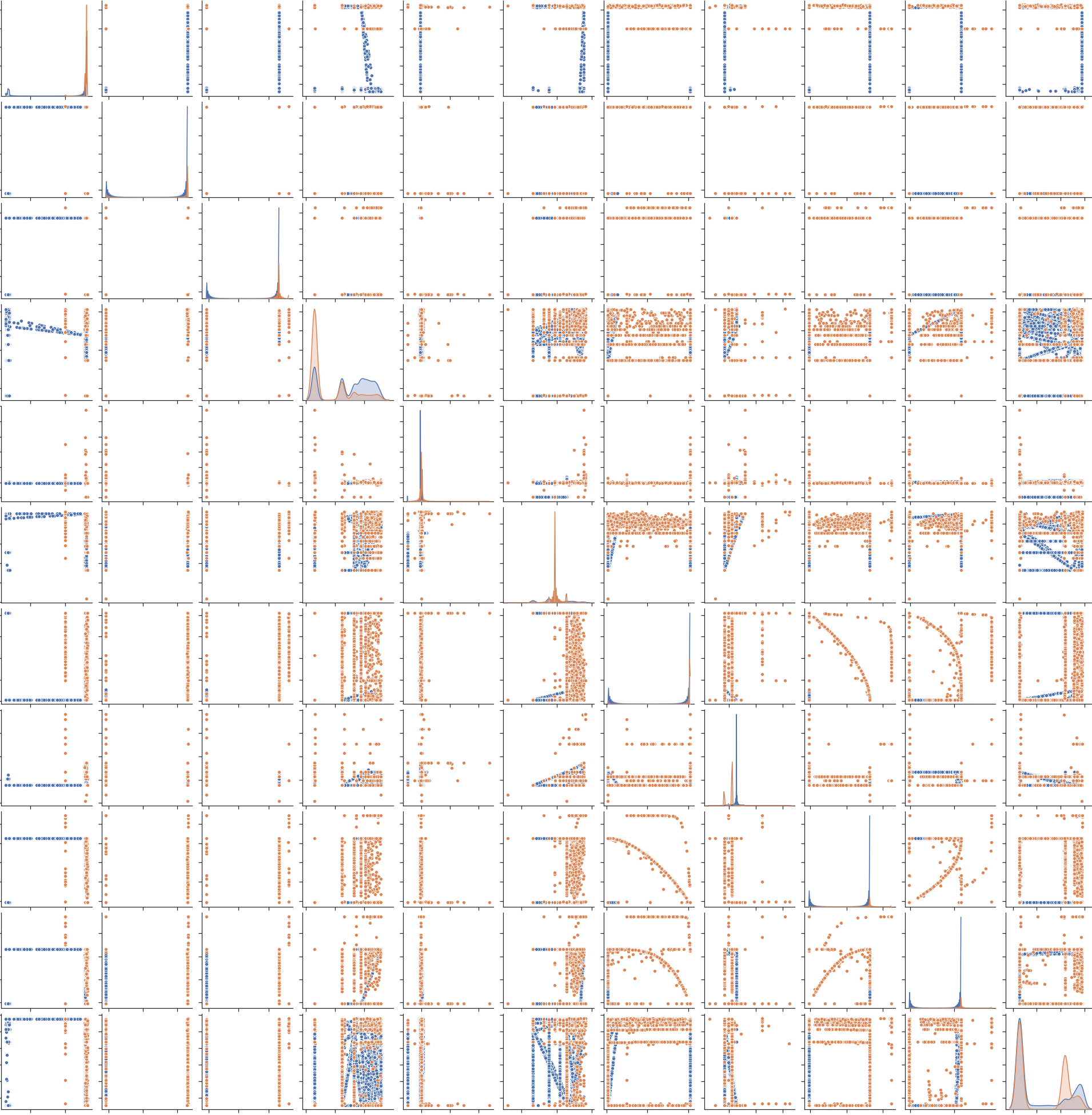}%
        \label{fig:PPADASYN}}%
    \caption{Pair plot of the dataset re-balanced with SMOTE and ADASYN algorithms.}
    \label{fig:PairPlotsADSM}
\end{figure*}

The pair plots for the original unbalanced MTA-KDD dataset, as well as for the G1No-balanced ones, have been already shown in Fig. \ref{fig:rrwOptMTAKDD} and \ref{fig:PPG1Nos}, respectively. The pair plots for the datasets obtained through SMOTE and ADASYN are shown in Fig. \ref{fig:PairPlotsADSM}.

The G1No algorithms mixed the distribution of the samples and changed the range of values, creating only sporadic small clusters and less outliers with respect to the original dataset. Moreover, it is worth noting that the spikes relative to the malware traffic have been preserved, whereas legitimate traffic has a Gaussian-like distribution skewed to the right. Moreover, boundary choices have been greatly reduced.


On the other hand, the standard deviation is rather low which makes us think that the mean (the highest point) is quite representative of many cases. However, the plots show some overlapping regions, as a consequence the difference in class densities becomes critical. This situation occurs because SMOTE and ADASYN generate small clusters with some features. This leads the densest class data points (usually the majority class) to cross regularly with the data point neighbours of the least dense class (usually the minority class) \cite{5196793}.


\subsection{Correlation matrix comparison}

\begin{figure*}[!ht] 
    \centering
    \subfigure[Correlation Matrix of G1No]{%
        \includegraphics[width=0.49\hsize]{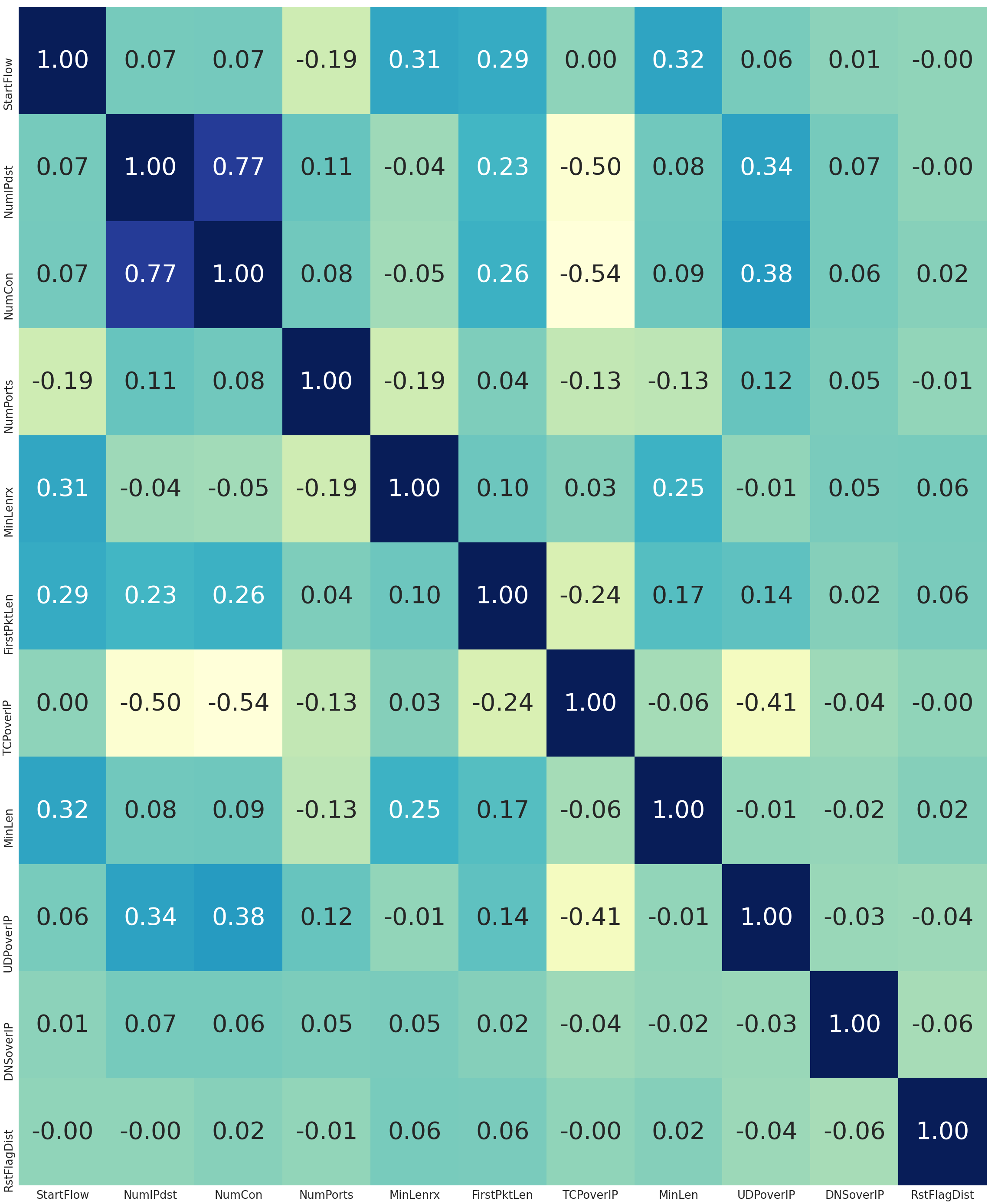}%
        \label{fig:CMG1No}}%
    \quad%
    \subfigure[Correlation Matrix of G1No Gourmet]{%
        \includegraphics[width=0.48\hsize]{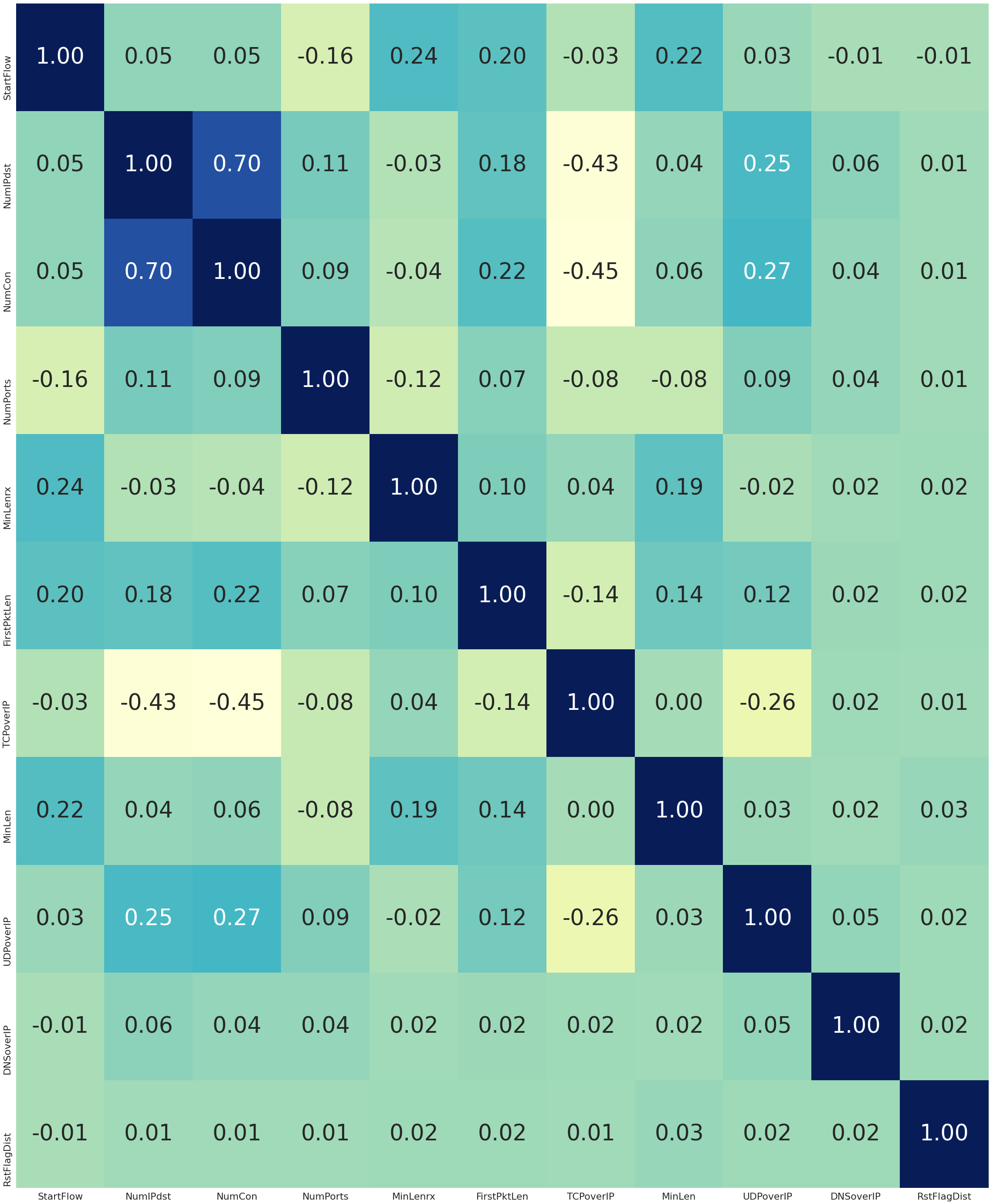}%
        \label{fig:CMG1Nog}}%
    \caption{Correlation Matrix of the MTA-KDD dataset rebalanced with the G1No algorithms.}
    \label{fig:corrmatrix}
\end{figure*}

\begin{figure*}[!ht] 
    \centering
    \subfigure[Correlation Matrix of SMOTE]{%
        \includegraphics[width=0.45\hsize]{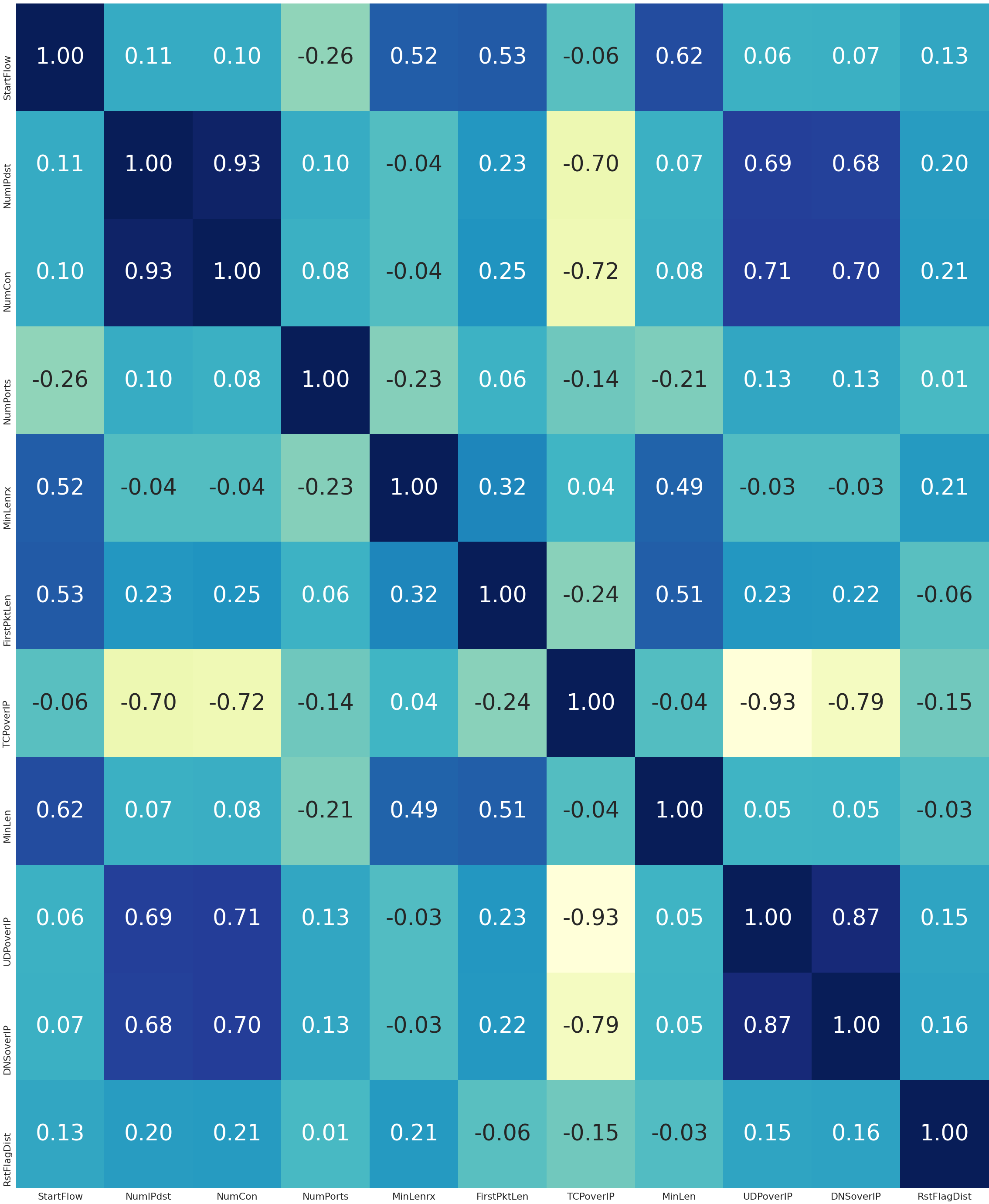}%
        \label{fig:CMSMOTE}}%
    \quad%
    \subfigure[Correlation Matrix of ADASYN]{%
        \includegraphics[width=0.45\hsize]{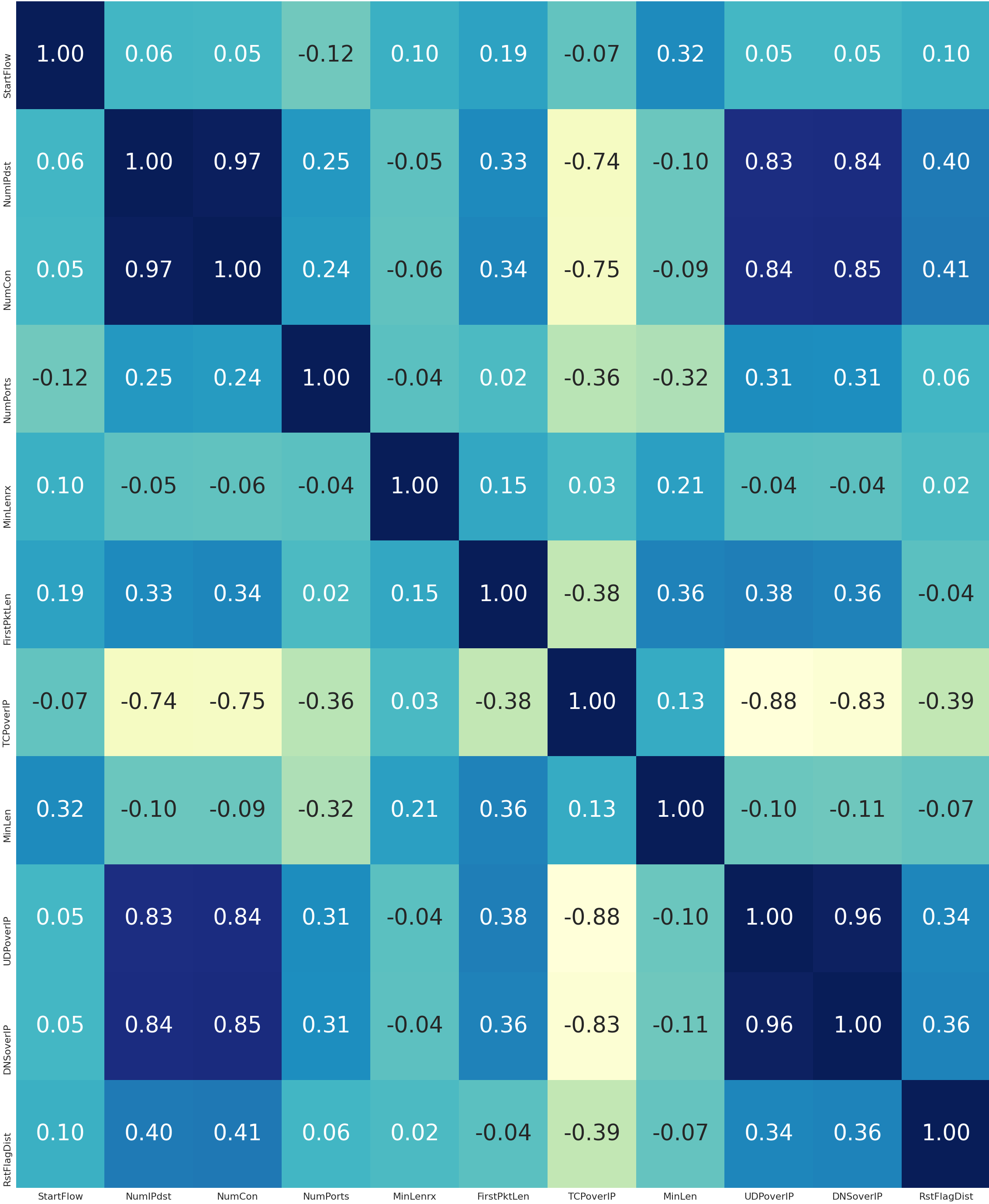}%
        \label{fig:CMADASYN}}%
    \caption{Correlation Matrix of the MTA-KDD dataset rebalanced with the SMOTE and ADASYN algorithms.}
    \label{fig:cmADSM}
\end{figure*}

Fig. \ref{fig:corrmatrix} shows the correlation matrix of the MTA-KDD dataset rebalanced with G1Nos algorithms, whereas Fig. \ref{fig:cmADSM} shows the matrix obtained from the same dataset rebalanced with SMOTE and ADASYN. In the matrices, each cell contains the Perason correlation coefficient between a pair of the 11 features in the MTA-KDD dataset, evaluated by comparing the available data samples.

The matrices in Fig. \ref{fig:corrmatrix} clearly show that with both the datasets generated by G1No, and in particular G1No Gourmet, all the features have a very low correlation respect to SMOTE and ADASYN. On the other hand, the balanced dataset generated by ADASYN and SMOTE have much more correlations feature close to 1, so there is a high chance that the performance of the classification model, applied to this dataset, will be impacted by the so called multicollinearity problem, which happens when the values of a  variable linearly predicted from the others with an high degree of accuracy \cite{Lieberman2014Mary}.

The different results may be due to the fact that the G1No methodology augments the minority class by adding samples with a Gaussian distribution, so that the generated samples inherit an approximate independence between features. In contrast, SMOTE-like oversampling uses linear combinations of the samples that amplify the correlation. 

Correlation is often interpreted as causation which is a big misconception. Correlation between variables does NOT indicate causation. Any highly correlated variable should be examined and thought of carefully. The balanced dataset generated by ADASYN and SMOTE have one feature close to 0.93, so there is a high chance that the performance of the model will be impacted by a problem called ``\textit{Multicollinearity}''. This can lead to skewed or misleading results.

It is interesting to remark that the G1No methodology augments the minority class adding samples that are by-construction sampled using a gaussian as the prior distribution. So generated samples inherit an approximate independence between features. In utterly contrast SMOTE-like oversampling use linear combinations of the samples so they exaggerate the problem of multicollinearity.


\subsection{MLP performance comparison}

Finally, we check the quality of the balanced datasets by measuring the classification performances reached by the same MLP emplyed in Section \ref{sect:silhDDI}.

Each dataset is split in an 85\% training set (59692 samples) and a 15\% \textit{test set} (10534 samples). Moreover, the training set is further split into a 15\% (8954 samples) \textit{validation set} and a 85\% \textit{learning set} (50738 samples). 

We start by analysing the \textit{Learning Curves} of the MLP, and in particular its loss curves. The Train Learning Curve (TLC), showing the loss evolution during the training phase of each epoch, tells us how well the model is learning, whereas the Validation Learning Curve (VLC), showing the loss evolution during the validation phase at the end of each epoch, tells us how well the model is generalizing.

\begin{figure*}[!ht] 
    \centering
    \subfigure[TLC and VLC of SMOTE]{%
        \includegraphics[width=0.48\hsize]{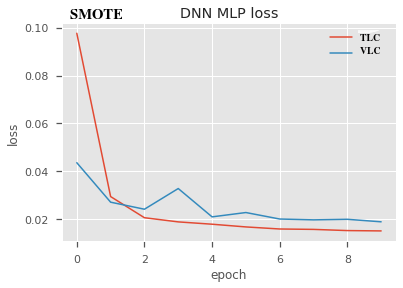}%
        \label{fig:LCSMOTE}}%
    \quad%
    \subfigure[TLC and VLC of ADASYN]{%
        \includegraphics[width=0.48\hsize]{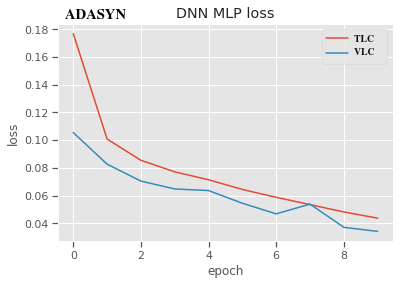}%
        \label{fig:LCADASYN}}%
    \quad%
    \subfigure[TLC and VLC of G1No]{%
        \includegraphics[width=0.48\hsize]{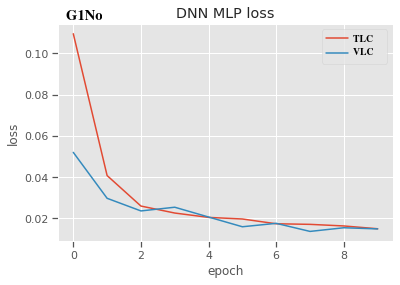}%
        \label{fig:LCG1No}}%
    \quad%
    \subfigure[TLC and VLC of G1No Gourmet]{%
        \includegraphics[width=0.48\hsize]{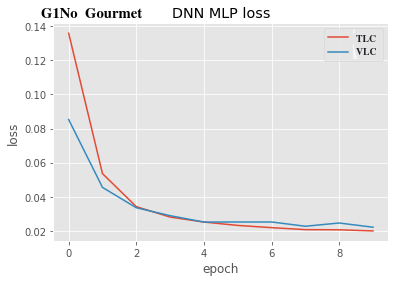}%
        \label{fig:LCG1NoG}}%
    \caption{TLC and VLC comparison on the four oversampling algorithms used.}
    \label{fig:TVlcROCG1No}
\end{figure*}

Fig. \ref{fig:LCSMOTE} shows that, with the dataset balanced with SMOTE algorithm, the validation loss is almost always higher than the training loss. This is called ``generalization gap'' and indicates that the model is not able to suitably generalize an it is overfitting.

Fig. \ref{fig:LCADASYN} shows that, with the dataset balanced with ADASYN algorithm, the training loss continues to decrease after the end of the plot. This indicates that the model is capable of further learning and that the training process was halted prematurely. Therefore, at the tenth epoch (which is our comparison epoch, as from the original experiments in \cite{itasec2020}), the model underfits.

On the other hand, Fig. \ref{fig:LCG1No} shows that, with the datatsets balanced with the two G1No algorithms, the training and validation loss quickly decrease to a point of stability with a minimal gap between the two final loss values. Thus the model, after the tench epoch, is suitably instructed and generalizing.

\begin{table}[!ht]
\centering 
\caption{Metrics on the testing set with the four balanced MTA-KDD datasets.}
\begin{tabular}{|l|c|c|c|r|}
\hline
&  \textbf{Precision} & \textbf{F1} & \textbf{AUC} & \textbf{Accuracy} \\ \hline
\textit{SMOTE}  & 99.52  & 99.50  & 99.90 & 99.49  \\ \hline
\textit{ADASYN} & 99.80  & 99.19  & 99.80  & 99.20 \\ \hline \hline

\textit{G1No} & \textbf{99.82}       & \textbf{99.69}   & \textbf{100.0}  & 99.69    \\ \hline
\textit{G1No Gourmet} & \textbf{99.84}       & \textbf{99.67}   & \textbf{100.0} & 99.70  \\ \hline
\end{tabular}
\label{tab:comparison}
\end{table}

Table \ref{tab:comparison} shows further metrics, measured at the tenth epoch, measured on the four validation sets, which confirm the superiority of the ones generated by the G1No algorithms. In particular, the G1No datasets outperform even the accuracy measured on the original MTA-KDD dataset, i.e., 99,60\% \cite{letteri2020dataset}. This may suggest that the silhouette-driven unbalancing technique described in Section \ref{sect:silhDDI}, followed by the G1No (possibly silhouette-driven, as for G1No Gourmet) re-balancing actually enhances the dataset, removing noisy data and allowing a more accurate training.

In general, while SMOTE and ADASYN generate sample points only on the faces and inside the convex simplex defined by the sample points, G1No generates samples inside and outside such simplex giving a richer structure, and this may be the key of its better performances.

\section{Conclusions}\label{sect:concl}
Our paper intends to increase a research interest on data difficulty factors that may deteriorate classifiers learnt from imbalanced data. Although most of the current research on class imbalance concerns the development of new algorithms, we claim that it is still worth to study the nature of imbalanced data, characteristics of the minority class distribution and its influence on classification performance. 
 
In this work, we elaborate a silhouette-driven methodology alone on the dataset used for the experiments and it has given promising results unbalancing the dataset until an ID of 0.3721 with Average Precision-Recall of 0.72. We can lighten the dataset considerably via undersampling increasing the performance because we are balancing the dataset but our G1nos algorithms are what really shines in the paper. 

We described our algorithms, called G1No and G1No Gourmet, which use a 1NN approach to a novel classification problem in sample generation with unbalanced class distribution. Specifically, we empirically studied the effects of the silhouette coefficient in undersampling dataset. Following them, we have introduced an identification method based on the analysis of sample silhouette coefficient class. We have shown a way of constructing the neighbourhood 1-nearest examples. Giving minority example allows us to identify a type of this example. 

The experiments show that the G1Nos algorithms surpass the major state of the art balancing techniques. We have experimentally shown that both these methods lead to similar results with 99.69\% and 99.70\% of accuracy but work better respect to the other two with more than half percentage point less. G1Nos algorithms show to us the promising character of a true augmentation technique as we will deepen in further work. Considering the Silhouette metric, we are going to consider a new idea about possible data transformation (e.g., PCA or better NCA) which can improve G1No Gourmet algorithm. 

During the evaluation of experiments, we always checked the correlation between different variables in the overall dataset and gather some insights as part of the exploration and analysis. We noted the importance, in this context, to analyse the correlations among the new instances in the features. They reveal to be very useful to determine differences among the four algorithms under observation. 

Finally, the issue of imbalance dataset raise many ethical questions and concerns related to the fairness of the classifiers trained using such datasets. Furthermore, in critical contexts where decisions of the ML algorithms might have ethical impacts (e.g. cause harm to humans), explainability of such decisions is fundamental \cite{Dyoub2020Letteri}.  These issues are subjects of our future works.

\bibliographystyle{splncs04}
\bibliography{biblio}
\end{document}